\documentclass[11pt,a4paper]{article}
\usepackage{times,latexsym}
\usepackage{url}
\usepackage[T1]{fontenc}
\usepackage[utf8]{inputenc}
\usepackage{microtype}
\usepackage{xcolor}
\usepackage{graphicx}
\usepackage{booktabs}
\usepackage{tabularx}
\usepackage{multirow}
\usepackage{multicol}
\usepackage{amsmath}
\usepackage{utfsym}
\usepackage{comment}
\usepackage{gb4e}
\noautomath

\usepackage[acceptedWithA]{tacl2021v1}

\title{Scope Ambiguities in Large Language Models}

 \author{
   Gaurav Kamath$^{\alpha, \beta}$\;
   Sebastian Schuster$^\gamma$\;
   Sowmya Vajjala$^\delta$\;
   Siva Reddy$^{\alpha, \beta, \epsilon}$
   \\
   \ \\
   $^\alpha$McGill University, 
   $^\beta$Mila - Quebec AI Institute,
   $^\gamma$University College London, 
   \\
   $^\delta$National Research Council Canada,
   $^\epsilon$Facebook CIFAR AI Chair
   \\
   \texttt{gaurav.kamath@mail.mcgill.ca}\;
    \texttt{s.schuster@ucl.ac.uk}\;
   \\ 
   \texttt{sowmya.vajjala@nrc-cnrc.gc.ca}\;
   \texttt{siva.reddy@mila.quebec}
 }

\date{}

\begin{document}
\maketitle
\begin{abstract}
Sentences containing multiple semantic operators with overlapping scope often create ambiguities in interpretation, known as \textit{scope ambiguities}.
These ambiguities offer rich insights into the interaction between semantic structure and world knowledge in language processing.
Despite this, there has been little research into how modern large language models treat them.
In this paper, we investigate how different versions of certain autoregressive language models\textemdash GPT-2, {GPT-3/3.5}, Llama 2 and GPT-4\textemdash treat scope ambiguous sentences, and compare this with human judgments.
We introduce novel datasets that contain a joint total of almost 1,000 unique scope-ambiguous sentences, containing interactions between a range of semantic operators, and annotated for human judgments.
Using these datasets, we find evidence that several models (i) are sensitive to the meaning ambiguity in these sentences, in a way that patterns well with human judgments, and (ii) can successfully identify human-preferred readings at a high level of accuracy (over 90\% in some cases).\footnote{Data and code are available at: \url{https://github.com/McGill-NLP/scope-ambiguity}}
\end{abstract}

\section{Introduction}
\label{sec:intro}

\begin{figure}
\includegraphics[width=\linewidth]{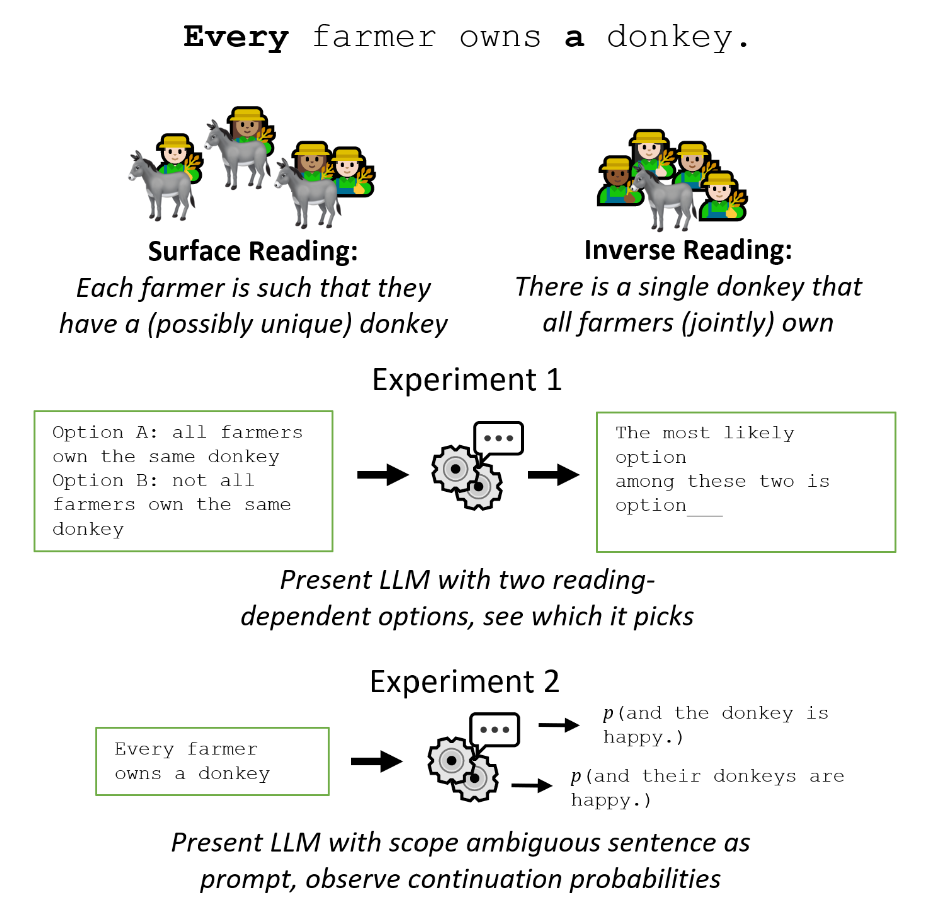}
\caption{A high-level overview of our study, showing our approaches to our first (see Section \ref{sec:experiment1a}) and second (see Section \ref{sec:exp2a}) experiments.}
\label{fig:pg1dg}
\end{figure}

Sentences like `every farmer owns a donkey' are systematically ambiguous between two readings: one in which the embedded noun phrase (NP) (e.g. `a donkey') is interpreted within the scope of the quantifier that precedes it (`every'), and another in which the embedded NP is interpreted outside its scope. As shown in Figure \ref{fig:pg1dg}, `every farmer owns a donkey', for example, could either mean (i) that each farmer simply owns their own (possibly unique) donkey, or (ii) that there is a specific donkey in question that all farmers jointly own.

Such constructions are examples of what are known as \textit{scope ambiguities}. They are called so because the standard account of these ambiguities is that they arise when the respective scope of multiple semantic operators in the expression is ambiguous, yielding more than one possible semantic structure.
Consider the following example:

\begin{exe}
\label{ex:farmer}
\ex 
\begin{xlist}
\ex \label{ex:farmersent} Every farmer owns a donkey.
\ex \label{ex:farmersurface}  Surface Scope: $\forall y[farmer(y) \rightarrow \exists x[donkey(x) \wedge owns(y,x)]]$
\ex \label{ex:farmerinverse}  Inverse Scope: $\exists x[donkey(x) \wedge \forall y[farmer(y) \rightarrow owns(y,x)]]$
\end{xlist}
\end{exe}

(\ref{ex:farmersent}), in logical form, involves a universal quantifier (introduced by `every'), and an existential quantifier (introduced by `a'). 
The ambiguity lies in the order of application (and thereby scopes) of these two operators. 
The surface scope reading of the sentence, (\ref{ex:farmersurface}), involves the universal quantifier outscoping the existential quantifier. 
The inverse scope reading, (\ref{ex:farmerinverse}), involves the reverse. 

Importantly for this present work, English speakers (i) have access to both kinds of readings, and (ii) generally disambiguate between them to arrive at a preferred reading \cite[see][]{kurtzman1993resolution}. 
For example, although (\ref{ex:farmersent}) has two possible interpretations, without further context, most people would prefer the surface reading, due to at least the surface positions of `a' and `every' in the sentence, as well as background world knowledge about farmers and donkeys \cite[see][for insights into how such factors affect reading preferences]{kurtzman1993resolution, saba2001plausible, anderson2004structure}. 

The focus of this paper is how large language models (LLMs) treat such ambiguities. 
Assessing how they do so offers important insights into interactions between semantic structure and world knowledge, as well as the representation of scope in LLMs:

\paragraph{Semantic Structure and World Knowledge} Scope disambiguation lies at the interface between natural language semantics and background world knowledge. 
Scope ambiguous sentences like (\ref{ex:farmersent}) are ambiguous between two semantic structures; disambiguating between these two possible structures (and the different readings they yield), however, often requires background world knowledge \cite{saba2001plausible}. 
Take the following two sentences:

\begin{exe}
\ex \label{ex:commonsense}
\begin{xlist}
    \ex \label{ex:commonsensesurface}Every conference attendee ate a Big Mac.
    \ex \label{ex:commonsenseinverse}Every conference attendee attended a networking event.
\end{xlist}
\end{exe}

Both examples in (\ref{ex:commonsense}) are scope-ambiguous in a similar way to (\ref{ex:farmer})\textemdash each offers two possible semantic structures yielding different readings.
However, choosing the preferred reading is easy in both cases: in (\ref{ex:commonsensesurface}), the surface scope reading (every attendee ate a potentially different Big Mac) is preferred, while in (\ref{ex:commonsenseinverse}), the inverse scope reading (there was a single networking event that all attendees attended) is preferred.
These preferences are a result of the general knowledge we have about conference attendees, networking events and Big Macs.

LLMs have been shown to be able to capture aspects of world knowledge \cite[e.g.,][]{roberts2020much, heinzerling2021language, alkhamissi2022review}, and, separately, to capture some properties of natural language semantics \cite[e.g.,][]{jawahar2019does, ettinger2020bert, pavlick2022}. 
Scope ambiguities present an opportunity to assess how they might integrate the two.

\paragraph{Scope Representation in LLMs} Model weights are largely uninterpretable, so despite generally high performance on language-based tasks, many questions remain about the abstract linguistic structures they capture \cite{belinkov2019analysis, hewitt2019structural, baroni2022proper}. 
The ambiguities discussed here arise out of a crucial component of linguistic structure: scope.
Analyzing how LLMs treat them helps us gain insight into how well they capture this component of structure.
This is particularly interesting because while formal logic, as in (\ref{ex:farmer}), allows for a straightforward, symbolic representation of scope ambiguities, it remains an open question whether vector-based LLM representations can adequately capture the multiple readings of such constructions. 

\paragraph{}
This paper therefore attempts to answer two questions:
\begin{itemize}
    \item[Q1:] \textit{Do LLMs exhibit similar preferences to humans in the interpretation of scope ambiguous sentences?}
    \item[Q2:] \textit{Are LLMs sensitive to the presence of more than one reading of scope ambiguous sentences?}
    
\end{itemize}

We conduct two experiments to investigate these questions.
From these experiments, we present evidence that the answer to these questions\textemdash at least for the more powerful models\textemdash is `yes'.

\section{Related Work}
\label{sec:relatedwork}

Scope ambiguities have been the focus in research within computational linguistics and natural language processing (NLP) primarily through the task of quantifier scope disambiguation, which involves the proper selection of a preferred scope reading given a scope-ambiguous sentence.

Early examples of NLP research on this task, such as \citet{higgins2003machine} and \citet{andrew2004statistical}, frame it as a classification task, and find models that outperform naive heuristics. 
Such work predates modern neural language models; \citet{rasmussen2022broad}, however, builds on this approach, framing quantifier scope disambiguation as a span-pair classification task.
They test RoBERTa \cite{liu2019roberta} on this task, and find that the model achieves higher accuracy on it than a majority-prediction baseline. 
This work, however, does not directly test the model's underlying linguistic capabilities; it tests the model on the classification task only after it is trained on examples from the dataset used.
As a result, it is unclear to what degree the model's performance on the test set is due to linguistic capabilities that emerged from its pretraining.

\citet{manshadi2011unrestricted} and \citet{tsiolis2020quantifier} approach the problem differently, as neither frames it as a classification task. 
\citet{manshadi2011unrestricted} represent scope relations as graphs, and frame the task as one of graph construction; they present a support vector machine that beats a naive heuristic baseline.
\citet{tsiolis2020quantifier}, on the other hand, attempts to reframe the task as a natural language inference task, Q\&A task, or one in which probabilities of continuations are compared.
They use a large language model\textemdash GPT-2 \cite{radford2019language} \textemdash but present mixed results.

Other research focuses on the linguistic factors that determine scope reading preferences in a corpus.
\citet{anderbois2012pragmatics} find that linear order, grammatical roles and lexical effects determine these preferences; \citet{leczkowski2022prepositions} build on this work and find that prepositions and preposition senses also affect scope reading preference.

The only two instances of work assessing how LLMs treat scope ambiguities in zero-shot contexts are, to our knowledge, recent works by \citet{Liu2023WereAL}, and \citet{stengel2023zero}.
The latter assesses how LLMs treat ambiguous inputs in terms of semantic parsing. 
The authors use templates to generate ambiguous sentences\textemdash including scope-ambiguous sentences\textemdash along with logical parses of them, and assess the abilities of LLMs to properly produce the two logical parses of each ambiguous sentence, in both few-shot and zero-shot contexts.
They find that models are poor at generating both parses of ambiguous sentences in zero-shot contexts, but can more accurately generate both parses in few-shot contexts.
\citet{Liu2023WereAL}, on the other hand, assess how LLMs treat linguistic ambiguity in terms of entailment relations.
Using prompting approaches to the task, as well as observing probabilities assigned to continuations of ambiguous sentences, they present evidence suggesting LLMs struggle to model ambiguity.

Both works, though they do not primarily focus on it, do include scope ambiguity data, and are thus relevant to our work. 
Where we diverge from these works, however, is in our data and experimental methods.
While the templates \citet{stengel2023zero} use allow for the generation of hundreds of sentences, they do limit the diversity of these stimuli; moreover, the scope ambiguities in their datasets are limited to instances of quantifier-quantifier interactions.
Similarly, \citet{Liu2023WereAL} estimate from a random sample that roughly 7.6\% of their data involves scope ambiguity; manually inspecting all 579 ambiguous sentences in their dataset, however, we find that the dataset contains a total of around 20 instances of scope ambiguity.
We also employ different experimental set-ups (see Sections \ref{sec:exp1method} and \ref{sec:exp2method}) than those used in the aforementioned works. 
Crucially, these experimental methods may be what provide us opposite findings from both of them; we discuss this difference in Section \ref{sec:conclusion}.

More broadly, our work belongs to a growing body of literature evaluating how well neural language models capture a range of semantic phenomena \cite[see][for an overview]{pavlick2022}.
This includes work assessing the capacity of such models to capture compositionality \cite[see e.g.,][]{ettinger2018assessing, shwartz2019, jawahar2019does, yu2020assessing, yu2021interplay} , as well as specific features such as negation \cite[see e.g.,][]{ettinger2018assessing,  kim2019probing, ettinger2020bert, jang2023consistency}, quantification \cite[e.g.,][]{kim2019probing, richardson2020probing, cui2022generalized}, and monotonicity \cite[e.g.,][]{yanaka2019can, yanaka2020neural, wijnholds2023assessing}.

\section{Background}
\label{sec:background}

\begingroup
\begin{table*}
\centering
\resizebox{\textwidth}{!}{
    \begin{tabular}{@{}lccc@{}}
    \toprule
   \multirow{2}{*}{\textbf{Interaction Type}} & \multirow{2}{*}{\textbf{Example}} & \multicolumn{2}{c}{\textbf{Count}} \\\cmidrule(lr){3-4}
   & & {Experiment 1} & {Experiment 2} \\\midrule
   \multirow{2}{*}{\textit{Quantifier-Quantifier}} & \multirow{2}{*}{\texttt{Every laptop is facing a glitch}} & 136/153 (88.9\%) & 7/29 (24.1\%) \\ & & \textbf{653/837 (78\%)*} & \textbf{38/110 (34.5\%)} \\ \midrule
   \multirow{2}{*}{\textit{Quantifier-Negation}} & \multirow{2}{*}{\texttt{I didn't pass all of my exams}} & 11/153 (7.2\%) & 11/29 (37.9\%) \\ & & \textbf{184/837 (22\%)} & \textbf{35/110 (31.8\%)} \\ \midrule
   \multirow{2}{*}{\textit{Quantifier-Adverb}} & \multirow{2}{*}{\texttt{I generally spar with two boxers}} & - & 11/29 (37.9\%) \\ & & - & \textbf{37/110 (33.6\%)} \\ \midrule
   \multirow{2}{*}{\textit{Quantifier-Misc.}} & \multirow{2}{*}{\texttt{Each truck is either green or red (but not both)}} & 6/153 (3.9\%) & - \\ & & - & - \\ 
   \bottomrule
    \end{tabular}
    }
     \caption{Original Experiment 1A and 2A datasets (regular), as well as expanded versions used in Experiments 1B and 2B (bold), broken down by interaction type. Our original Experiment 1A dataset consisted of a few examples involving disjunction, as shown in the example above. We label these as a miscellaneous type of interaction, as they are not our primary focus. *These include a balanced set of `quantifiers', including numbers, indefinites and quantificational determiners.}
     \label{tbl:datasetbreakup}
\end{table*}
\endgroup

We focus on scope ambiguities involving quantifiers such as `\textit{some}', `\textit{every}' and `\textit{most}', as well as quantifier-like determiners, like indefinites and numbers. 
(\ref{ex:farmer}) is an example of scope ambiguity arising out of quantifier-quantifier interactions.
But scope ambiguities can also arise out of quantifier-negation and quantifier-adverb interactions, as shown below:

\noindent{\underline{\textit{Quantifier + Negation:}}}
\begin{exe}
    \ex Sita \textbf{doesn't} like \textbf{a} classmate of hers.
    \label{ex:qneg}
    \ex 
    \begin{xlist}
        \ex Surface Reading: There is no classmate that Sita likes. 
        \label{ex:qnegsurface}
        \ex Inverse Reading: There is a specific classmate that Sita does not like. 
        \label{ex:qneginv}
    \end{xlist}
\end{exe}

\noindent{\underline{\textit{Quantifier + Adverb:}}}
\begin{exe}
    \ex Bachi \textbf{usually} meets \textbf{two} professors.
    \label{ex:qadv}
    \ex 
    \begin{xlist}
        \ex Surface Reading: Usually, Bachi meets any two professors, who are possibly different each time.
        \label{ex:qadvsurface}
        \ex Inverse Reading: There are two professors who Bachi meets regularly.
        \label{ex:qadvinv}
    \end{xlist}
\end{exe}

In all three cases, the different readings have different truth conditions, and each is therefore logically compatible with a different set of propositions. 
As an illustration, consider our original example, reproduced here as (\ref{ex:restrictionssent}):

\begin{exe}
    \ex Every farmer owns a donkey.
    \label{ex:restrictionssent}
    \ex 
    \begin{xlist}
        \ex Each farmer has a different donkey.
        \label{ex:restrictionssurface}
        \ex All farmers have the same donkey.
        \label{ex:restrictionsinverse}
    \end{xlist}
\end{exe}

(\ref{ex:restrictionssurface}) is logically compatible with (\ref{ex:restrictionssent}) only given the surface scope reading of the sentence, which states that each farmer has a potentially unique donkey. 
It is not logically compatible with the inverse scope reading of the sentence, which states that there is an individual donkey that all farmers (jointly) have.
(\ref{ex:restrictionsinverse}), however, is also logically compatible with the inverse scope reading of the sentence.
In Experiments 1A and 1B, we use these differences in logical compatibility to assess whether LLMs exhibit similar preferences to humans in the interpretation of scope ambiguous sentences.

Similarly, different scope readings often yield different effects in a discourse setting. 
Consider (\ref{ex:qadv}): under the inverse scope reading, two professors are introduced as constant across the instances of Bachi's meetings. 
Consequently, they can therefore be further referred to in the discourse, as in (\ref{ex:discourseinv}).

\begin{exe}
    \ex 
    \label{ex:discourse}
    \begin{xlist}
        \ex He likes those two professors.
        \label{ex:discourseinv}
        \ex It's a different pair each time.
        \label{ex:discoursesurface}
    \end{xlist}
\end{exe}

But under the surface scope reading of (\ref{ex:qadv}), there aren't necessarily two professors that are constant across instances, and who can therefore be further referred to in the discourse. 
As a result, (\ref{ex:discourseinv}) is not an acceptable follow-up.
The possible variability of the professors across multiple instances, however, does mean that (\ref{ex:discoursesurface}) is an acceptable follow-up (which it is not given the inverse scope reading).
In Experiments 2A and 2B, we use such patterns of acceptable and unacceptable followups to assess whether LLMs are sensitive to the presence of multiple readings of scope ambiguous sentences.

\section{Experiment 1A}
\label{sec:experiment1a}

\begingroup
\begin{figure*}
\small
\setlength\tabcolsep{1em}
\renewcommand{\arraystretch}{1.5}
\centering
    \begin{tabular}{|p{15cm}|}
    \hline
   \textbf{\texttt{There are exactly six chairs evenly spaced around a circular table.}} \\
    \texttt{On the basis of this phrase/statement alone, and with no further context, there are two options:} \\
    \textbf{\texttt{Option A: the six chairs are all around the same table}} \\
    \textbf{\texttt{Option B: the six chairs aren't all around the same table}} \\
    \texttt{Specifically in relation to this context, \textcolor{orange}{[the most likely option among these two is option]} \textcolor{cyan}{[which of these two options is most likely?]}} \\
    \hline
    \end{tabular}
     \caption{An example of stimuli provided to models in Experiments 1A and 1B. The sections highlighted in bold are taken from our Experiment 1A dataset, and vary between individual stimuli presented to the models. The non-highlighted sections, which act as a prompt frame, remain fixed. For chat-optimized models, we solicit the model's response using the question highlighted in blue; for plain autoregressive models, we solicit the model's response by seeing what it predicts after the sequence highlighted in orange. In the control setting, the ambiguous sentence is dropped.}
     \label{fig:exp1dataset}
\end{figure*}
\endgroup

\subsection{Method}
\label{sec:exp1method}

In our first experiment, we assess whether LLMs show similar preferences to humans in how scope ambiguous sentences are interpreted. 
We frame this as a Q\&A task. 
We present the LLM with sentences that are technically scope ambiguous, but have one strongly preferred scope reading (in some cases, this is a surface scope reading, and in others, it is an inverse scope reading).  
We then present the model with two possible statements based on this ambiguous sentence.
One statement is compatible with only the surface scope reading of the ambiguous sentence, while the other statement is compatible with the inverse scope reading. 
In the case of chat-optimized models, we then ask the model which option is more likely; in the case of models not optimized for chat, we then obtain this answer through next token prediction.
Finally, we observe whether these responses align with the reading preferred by most humans.

Figure \ref{fig:exp1dataset} shows an example of how we conduct this method, using the dataset we develop for this experiment (details in Section \ref{sec:exp1dataset}). 
We concatenate, with newlines as shown in Figure \ref{fig:exp1dataset}, (i) a test sentence that is technically scope-ambiguous; (ii) an explanation that there are two options; (iii) two statements, labelled Option A and Option B, where one is compatible only with the surface scope reading, and the other is compatible with the inverse scope reading; and (iv) a prompt that elicits the model's preferred choice. 
In the case of chat-optimized models, we observe the model's response to a question asking it to choose between the options.
For other models, we observe the next token predicted by the model after the text `\texttt{the most likely option among these two is option}': this is either `A' or `B', corresponding to Option A and Option B respectively. 
We treat these as the model's `answer', and evaluate the model based on whether it aligns with preferred human readings.

\subsubsection*{\textit{Control: No Sentence in Prompt}}
One possible issue with this approach is that answers may depend more on the likeliness of the two options as general statements than on their likeliness \textit{given} the ambiguous sentence. 
We therefore add a control: we remove the ambiguous sentence altogether from the stimulus, and present the rest of it to the model, conducting the same task as before. 
In this setting, the model should do significantly worse, as it is not exposed to the sentence that is being evaluated with respect to the two options.
For instance, in the example in Figure \ref{fig:exp1dataset}, both options appear plausible in the absence of any context; following the scope-ambiguous sentence, however, only Option A is plausible.
If model performance does not drop significantly when the ambiguous sentence is dropped, the model's performance in the original setting is likely unrelated to its processing of the ambiguous sentence, and instead reflects the background likeliness of each option.

\subsection{Dataset}
\label{sec:exp1dataset}

We build upon the quantifier scope dataset presented by \citet{anderbois2012pragmatics}.
We chose this dataset as a starting point because among the few existing scope ambiguity datasets (see Section \ref{sec:relatedwork}), it was the dataset that had datapoints most appropriate to the focus of this study.
This dataset contains around 1,700 sentences and phrases scraped from LSAT (Law School Admission Test) logic puzzles, and marked for quantifier scope where present.
We filtered the dataset for instances of two interacting `quantifiers'.\footnote{These included constructions such as `\textit{per}' in `\textit{one person per appointment}'\textemdash constructions that have some quantificational force, even if they aren't quantifiers in the strict sense\textemdash as well as instances of negation such as `\textit{none}' in `\textit{none of the cities}'.}
This narrowed it down to around 400 datapoints.
Next, we manually constructed contrasting `options' based on surface and inverse scope readings for whichever of these roughly 400 datapoints allowed this approach, giving us 186 sentences with accompanying contrasting statements.

To further ensure that these datapoints had strong scope reading preferences, we then conducted two rounds of human validation. 
In both rounds, we recruited participants via Prolific.
Participants (38 in each round) were presented the filtered scope-ambiguous sentences along with two accompanying options, and were asked to pick the most likely option.
In the first round, we reworded datapoints with low subject agreement; in the second round, we dropped any datapoints with less than 75\% agreement (all datapoints received at least 4 evaluations).
For the datapoints that remained, gold labels (i.e. the correct option for each datapoint, and consequently, the preferred scope reading) were taken as the majority vote from study participants.
This process ultimately yielded 153 scope ambiguous sentences, each with a pair of options.

Of these, 41 had an inverse scope reading preferred, while the remaining 112 had a surface scope reading preferred.
Almost all were examples of scope ambiguities arising from quantifier-quantifier interactions, with a handful involving quantifier-negation interactions, and even fewer involving other types of interactions (see Table \ref{tbl:datasetbreakup} for a breakdown by interaction type). 
As a final step, we duplicated each datapoint, but with a flipped order of options (i.e. `Option A' was labelled `Option B', and vice versa).
This meant that while the distribution of preferred scope-readings remained skewed, the final dataset\textemdash which contains 306 datapoints covering 153 unique sentences\textemdash had an even distribution of correct answers (50\% `A' and 50\% `B').

\subsection{Models}
\label{sec:exp1models}

For all experiments, we choose to use autoregressive language models, due to their growing prevalence in practical applications using prompting.

Specifically, for this experiment, we use chat and vanilla versions of Llama 2 \cite{touvron2023llama} at 7B, 13B and 70B sizes\footnote{For Llama 2 at 70B, we use a version of the model loaded in 8-bit (see \citet{dettmers2022llm}).}, three variants of GPT-3/3.5 \cite{brown2020language, ouyang2022training}: \texttt{davinci}, \texttt{text-davinci-002}, and \texttt{text-davinci-003}, and GPT-4 \cite{openai2023gpt4}. 
See Table \ref{tbl:modeltable} for a summary of key differences between these models.

\begin{table}
    \centering
    \tiny
    \begin{tabular}{@{}m{0.4\columnwidth}p{0.4\columnwidth}c@{}}
         \toprule
    \textbf{Model} & \multicolumn{1}{l}{\textbf{Size(s)}} & \multicolumn{1}{c}{\textbf{RLHF?}} \\ \midrule
    {GPT-3-davinci} & {175B} & {\usym{274C}} \\
    {GPT-3.5-text-davinci-002} & {Unclear} & {\usym{2731}} \\
    {GPT-3.5-text-davinci-003} & {Unclear} & {\usym{1F5F8}} \\
    {GPT-3.5-turbo} & {Unclear} & {\usym{1F5F8}} \\
    {GPT-4} & {Unclear; possibly ensemble model} & {\usym{1F5F8}} \\
    {Llama 2} & {7B, 13B, 70B} & {\usym{274C}} \\
    {Llama 2 Chat} & {7B, 13B, 70B} & {\usym{1F5F8}} \\
    {GPT-2} & {117M, 345M, 774M, 1.5B} & {\usym{274C}} \\
    \end{tabular}
    \caption{Summaries of models used for Experiments 1 and 2, including size and whether they were fine-tuned with reinforcement learning from human feedback (RLHF). *\texttt{text-davinci-002} is not fine-tuned with RLHF, but is fine-tuned on human demonstrations and highly rated model outputs.}
    \label{tbl:modeltable}
\end{table}

\subsection{Human Baselines}
\label{sec:exp1baselines}

After the human feedback-based filtering mentioned above, we conducted another round of the same experiment with humans to get human baselines on this dataset.
We are testing models on their ability to choose the scope readings preferred by most people\textemdash but how good are human themselves at choosing the scope readings preferred by most other people?
Our human baselines should provide a sense of the answer to this question.
68 native speakers of English were recruited via Prolific for a repeat of the experimental set-up described in Section \ref{sec:exp1dataset}, but this time with the final dataset. 
Each participant was presented with 18 datapoints and evaluated on their answers. 
We then calculated overall accuracy as the total proportion of correct responses out of all responses. 

\subsection{Results}
\label{sec:exp1results}

\begin{table}
\centering
\resizebox{\columnwidth}{!}{
\begin{tabular}{@{}lcccccc@{}}
    \toprule
    \multirow{2}{*}{\textbf{Source}} & \multicolumn{2}{c}{\textbf{Accuracy}} & \multicolumn{2}{c}{\textbf{Surface Acc.}} & \multicolumn{2}{c}{\textbf{Inverse Acc.}} \\\cmidrule(lr){2-3}\cmidrule(lr){4-5}\cmidrule(lr){6-7}
    & \textit{test} & \textit{control} & \textit{test} & \textit{control} & \textit{test} & \textit{control} \\\midrule
    Human* & 0.90 & - & 0.89 & - & 0.91 & - \\
    Llama2-7b & 0.58 & 0.67 & 0.57 & 0.69 & 0.61 & 0.63 \\
    Llama2-7b-chat & 0.54 & 0.55 & 0.54 & 0.58 & 0.54 & 0.49 \\
    Llama2-13b & 0.71 & 0.63 & 0.73 & 0.65 & 0.65 & 0.59 \\
    Llama2-13b-chat & 0.72 & 0.67 & 0.74 & 0.69 & 0.67 & 0.61 \\
    Llama2-70b & 0.88 & 0.70 & 0.88 & 0.71 & 0.87 & 0.68 \\
    Llama2-70b-chat & 0.85 & 0.71 & 0.88 & 0.72 & 0.78 & 0.67 \\
    GPT-3-davinci & 0.58 & 0.58 & 0.58 & 0.58 & 0.57 & 0.57 \\
    GPT-3.5-td002 & 0.80 & 0.72 & 0.83 & 0.72 & 0.73 & 0.71 \\
    GPT-3.5-td003 & 0.91 & 0.75 & 0.94 & 0.79 & 0.84 & 0.67 \\
    GPT-3.5-turbo & 0.80 & 0.67 & 0.81 & 0.66 & 0.77 & 0.66 \\
    \textbf{GPT-4} & \textbf{0.98} & \textbf{0.75} & \textbf{0.98} & \textbf{0.79} & \textbf{0.99} & \textbf{0.65} \\
    \bottomrule
\end{tabular} 
}
\caption{Results from Experiment 1A: model accuracy, as well as accuracy on sentences that had a preferred surface or inverse reading. In the test setting, the ambiguous sentence is present in the prompt; in the control setting it is dropped. *Values for humans are averaged across all participants' responses.}
\label{tbl:exp1results}
\end{table}

The results of Experiment 1A are shown in Table \ref{tbl:exp1results}. 
Human responses yield an average accuracy of around 90\%, suggesting that English speakers can, with a high degree of accuracy, arrive at scope reading preferences shared by most other people.

When it comes to model responses, although models like \texttt{davinci} and \texttt{Llama2-7b} do not perform far above chance (50\%, since correct answers in the dataset were balanced through duplication), several other models do achieve high performance\textemdash both versions of \texttt{Llama2-70b}, as well as all the GPT-3.5 models achieve an accuracy of 80\% or more in the test setting, while GPT-4 achieves 98\%, close to the ceiling.

The control setting adds further insights to these results. 
Neither \texttt{davinci} nor the versions of Llama 2 at 7B see their accuracy scores drop when prompts are provided without the actual scope ambiguous sentence (\texttt{Llama2-7b} actually produces a higher accuracy in this setting), suggesting that performance in the test setting is not a result of the models' processing of the ambiguous input, but primarily driven by the background likeliness of the two options. The models with higher accuracy scores, however, see more severe drop-offs in the control setting, most notably with GPT-4, which sees its accuracy drop to 75\% in the control setting. 

\subsection{Discussion}
\label{sec:exp1discussion}

These results suggest that the more advanced LLMs evaluated\textemdash GPT-3.5, Llama 2 at 70B, and most notably GPT-4\textemdash are able to exhibit similar scope reading preferences as humans, with a high level of accuracy.
Smaller or less advanced models, however, such as Llama 2 at 7B, appear to fail. 

Also worth noting is the fact that, for almost all models, performance on sentences that had a preferred inverse scope reading was lower than on those that had a preferred surface scope reading.
This aligns with literature suggesting that inverse scope readings are generally harder to access than surface readings \cite[see e.g.,][]{kurtzman1993resolution, anderbois2012pragmatics}, but curiously, does not align with the behaviour of humans in this experiment, who showed no such dispreference.

The deeper implication of some of the models' high performance, however, is that LLMs can not only capture different types of readings\textemdash surface and inverse, which correspond to different semantic structures\textemdash but also integrate background world knowledge in their behavioural preferences when confronted with scope ambiguous constructions.

\section{Experiment 2A}
\label{sec:exp2a}

\subsection{Method}
\label{sec:exp2method}

\begingroup
\begin{figure*}
\setlength\tabcolsep{1em}
\renewcommand{\arraystretch}{1.5}
\centering
    \includegraphics[width=\linewidth]{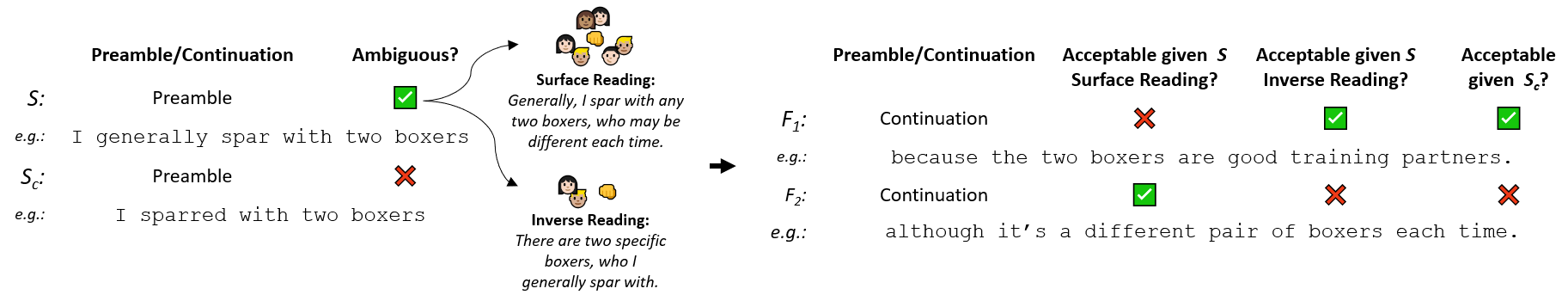}
    \caption{Experiment 2A and 2B set-up, comprising of an ambiguous sentence $S$, unambiguous control $S_{c}$, and two follow-ups, $F_{1}$ and $F_{2}$, demonstrated using an example from our manually constructed dataset. We compare the probabilities a model assigns to $F_{1}$ and $F_{2}$ as continuations to $S$, versus as continuations to $S_{c}$.}
     \label{fig:exp2dataset}
\end{figure*}
\endgroup

Our first experiment showed us that, where there is a clear preferred reading, LLMs can mimic human preferences in the interpretation of scope ambiguous sentences. 
It did not, however, indicate whether or not LLMs were sensitive to the fact that each such scope ambiguous sentence had more than one reading.
This question is the focus of our second experiment.

Here, we assess whether models exhibit different behaviour for scope ambiguous sentences than they do for similar, \textit{non} scope ambiguous sentences, in a manner that indicates a sensitivity to the meaning ambiguity in the former but not the latter.
We do not frame this as a conventional goal-oriented task such as Q\&A.\footnote{This was mainly due to our focus on examples without one strongly preferred reading (see Section \ref{sec:exp2dataset}). Without one strongly preferred reading, tasks like Q\&A, which require a `right' and `wrong' answer, are difficult to implement.} 
Instead, following work that brings psycholinguistic methods to language model analysis \cite[see][]{linzen2016assessing, futrell-etal-2019-neural, ettinger2020bert, baroni2022proper,schuster-linzen-2022-sentence}, we investigate the question by observing the probabilities a model assigns to different types of continuations given a scope-ambiguous sentence.

Figures \ref{fig:pg1dg} and \ref{fig:exp2dataset} illustrate the general set-up we employ:
we begin by presenting the LLM with a scope ambiguous sentence $S$. 
We then observe the probabilities the model assigns to two followups to $S$, labelled $F_{1}$ and $F_{2}$.
$F_{1}$ is an acceptable continuation to $S$ only given the inverse scope reading of $S$, while $F_{2}$ is an acceptable continuation to $S$ only given the surface scope reading of $S$.
We then compare these probabilities with those the model assigns to $F_{1}$ and $F_{2}$ given a control sentence, $S_{c}$.
$S_{c}$ is highly similar in both syntax and semantics to $S$, but differs in that it is not scope ambiguous; $F_{1}$ remains an acceptable continuation to $S_{c}$, though $F_{2}$ does not.
 
If an LLM successfully captures the meaning ambiguity of a sentence like $S$, we would expect the ratio of probabilities it assigns to $F_{1}$ and $F_{2}$ as continuations to $S$ to be smaller than the ratio of probabilities it assigns to the same followups as continuations to $S_{c}$.
In other words, we expect the following inequality to hold:

\begin{exe}
\ex \label{ex:ineq} \Large
$\frac{P(F_{1}|S)}{P(F_{2}|S)} < \frac{P(F_{1}|S_{c})}{P(F_{2}|S_{c})}$
\end{exe}

This is because while $S$ is ambiguous between two readings, and therefore allows for both $F_{1}$ and $F_{2}$ as continuations, $S_{c}$ has only one reading, and allows only for $F_{1}$ as a continuation (see Figure \ref{fig:exp2dataset}).
If an LLM upholds (\ref{ex:ineq}), it thus provides evidence of capturing the fact that $S$ is ambiguous between two readings, in a way that non scope-ambiguous sentences, even if syntactically and semantically similar, are not.

For a more thorough analysis, we also observe the degree to which ${P(F_{1}|S_{c}):P(F_{2}|S_{c})}$ is greater than ${P(F_{1}|S):P(F_{2}|S)}$. 
We measure the difference in the log ratios of $F_{1}$ and $F_{2}$ given $S$, and $F_{1}$ and $F_{2}$ given $S_{c}$, and use it to calculate what we call the model's ambiguity recognition score, or `$\alpha$-score':

\begin{exe}
\ex \label{ex:ambrecformula}
$\alpha = $ $ - [[log P(F_{1}|S) - log P(F_{2}|S)] $ \\ $ - [log P(F_{1}|S_{c}) - log P(F_{2}|S_{c})]]$
\end{exe}

If the inequality in (\ref{ex:ineq}) holds, the $\alpha$-score will be positive; the larger its value, the greater the difference in ratios.

\subsection{Dataset}
\label{sec:exp2dataset}

Existing scope ambiguity datasets (see Section \ref{sec:relatedwork}) are (i) few in number, and (ii) generally involve examples where one scope reading is strongly preferred over the other.
While we made use of this second observation in our first experiment, where we assessed whether LLMs exhibited similar scope reading preferences as humans, it is a significant problem for the current experiment, due to its aims.

In this experiment we aim to test LLMs for their sensitivity to the presence of multiple readings of a scope ambiguous sentence.
But in such cases, if humans themselves find one of these readings very hard to access without further context, it would be unfair to expect models to do so.
For a fair evaluation, therefore, it is crucial that we use sentences which do not have one reading strongly preferred over another.

We therefore construct a small-scale dataset, consisting of 38 manually handcrafted datapoints, where each datapoint includes a scope ambiguous sentence ($S$), a matching non scope-ambiguous control sentence ($S_{c}$), and two follow-up phrases ($F_{1}$ and $F_{2}$), yielding a total of 152 sentence-continuation pairs. 
For further validation, these datapoints were then filtered through our human baselines: any datapoints that yielded negative human-derived scores (details in Section \ref{sec:exp2baselines}) were dropped, as such scores indicated that these were datapoints for which the inequality in (\ref{ex:ineq}) did not align with human judgments.
This left us with 29 unique datapoints, yielding 116 unique sentence-continuation pairs (see Table \ref{tbl:datasetbreakup} for a breakdown by interaction type).

\subsection{Human Baselines}
\label{sec:exp2baselines}

Since the current experiment involves the analysis of probabilities assigned to text sequences\textemdash something not directly replicable with humans\textemdash we use proxy scores derived from a crowdsourced judgment task as our human baselines.
We conducted a crowdsourced study via Prolific, involving 140 native speakers of English; each was presented random sentence-continuation pairs from the dataset, and asked to provide ratings from 1 to 7 on how `natural-sounding' the continuation was to the sentence.
From these ratings, we computed the mean score for each sentence-continuation pair, and normalized them to be in an interval between 0 and 1.
We then treat these normalized scores as we treat model-assigned probabilities when calculating $\alpha$-scores 
and label the negative difference of log ratios our proxy $\alpha$-score. 

This proxy score gives us an indirect means by which to compare human judgments of scope-ambiguous sentences and continuations with LLM-assigned probabilities of the latter given the former.
Just as in the case of $\alpha$-scores for models, we would expect human proxy scores to be positive.

\subsection{Models}
As in Experiment 1A, we work with autoregressive LLMs. 
Unlike in Experiment 1A, however, the current experimental set-up allows us to also work with models ill-suited to zero-shot contexts. 
We therefore ran this experiment not only on the models tested before, but also on several smaller variants of GPT-2 \cite{radford2019language}: small (117M params), medium (345M params), large (774M params) and XL (1.5B params).\footnote{Probabilities from GPT-2 and Llama 2 models were extracted using the \texttt{minicons} library \cite{misra2022minicons}.}
The reliance on probabilities, however, forces us to omit \texttt{GPT-3.5-turbo} and GPT-4, for which sequence log-probabilities are not accessible.

\subsection{Results}
\label{sec:exp2results}

\subsubsection{Mean Scores}
\label{sec:exp2resultsmeanscores}

We first compute mean $\alpha$ and proxy scores, along with $p$-values derived from paired $t$-tests.\footnote{We choose this statistical measure as both $\alpha$ and proxy scores are calculated as paired differences of differences.} 
Positive, statistically significant mean scores point to an overall sensitivity to the meaning ambiguity of the sentences in the dataset; comparing between models, higher mean scores also suggest stronger overall sensitivity.\footnote{While the latter is true between models, this relationship is less clear when comparing model scores and human scores. The main reason is that while human scores are derived from a bounded set of ratings between 1 and 7, model log-probabilities practically have no negative bound, allowing for more extreme differences between them. Since $\alpha$-scores are computed as differences of log differences (see (\ref{ex:ambrecformula})), it is thus possible for them to be much higher than derived human scores, purely on account of being unbounded below zero.\label{fnote:alphaexp}}
Table \ref{tbl:meansandcorr} shows our results.
All models yield positive mean $\alpha$-scores; and barring the case of \texttt{GPT-2-small}, all are statistically significant at a threshold of $p < 0.05$.

\begin{table}
\centering
\resizebox{\columnwidth}{!}{
\begin{tabular}{@{}lccccc@{}}
    \toprule
    \multirow{2}{*}{\textbf{Source}} & \multicolumn{2}{c}{\textbf{Mean Scores}} & \multicolumn{2}{c}{\textbf{Correlations}} & \multirow{2}{*}{\textbf{\boldmath$\alpha$ > 0}} \\\cmidrule(lr){2-3}\cmidrule(lr){4-5}
    & $\alpha$/proxy score & $p$-value & R-value &  $p$-value\\\midrule
    Human & 1.22 & 4.43e-06 & 1.0 & - & 1.0 \\ 
    GPT-2-small & 0.29 & 0.43 & 0.32 & 0.09 & 0.52 \\
    GPT-2-med & 0.97 & 0.03 & 0.38 & 0.04 & 0.62 \\ 
    GPT-2-large & 1.51 & 1.97e-03 & 0.33 & 0.08 & 0.69 \\ 
    GPT-2-xl & 1.79 & 9.78e-04 & 0.29 & 0.12 & 0.76 \\ 
    Llama2-7b & 3.89 & 2.9e-07 & 0.17 & 0.39 & 0.93 \\
    Llama2-7b-chat & 5.03 & 9.45e-07 & 0.27 & 0.16 & 0.86 \\
    Llama2-13b & 3.62 & 6.67e-07 & 0.49 & 7.38e-03 & 0.90 \\ 
    Llama2-13b-chat & 4.54 & 1.28e-05 & 0.53 & 3.31e-03 & 0.83 \\
    Llama2-70b & 3.97 & 5.74e-08 & 0.38 & 0.04 & 0.93 \\
    Llama2-70b-chat & 4.72 & 3.52e-06 & 0.43 & 0.02 & 0.90 \\
    GPT-3-davinci & 3.77 & 4.95e-08 & 0.20 & 0.31 & 0.93 \\
    GPT-3.5-td002 & 4.06 & 1.5e-04 & 0.41 & 0.03 & 0.83 \\
    \textbf{GPT-3.5-td003} & \textbf{8.36} & \textbf{1.44e-06} & \textbf{0.62} & \textbf{2.93e-04} & \textbf{1.0} \\
    \bottomrule
\end{tabular}
}
\caption{Results from Experiment 2A.
\underline{Mean Scores}: Mean $\alpha$ (for models) and proxy (for humans) scores with $p$-values from paired $t$-tests ($df = 28$).
\underline{Correlations}: Pearson correlation coefficients (R-values) between each model's $\alpha$ scores and human proxy scores, with derived $p$-values ($n = 29$).
\underline{$\alpha > 0$}: Proportion of datapoints where $\alpha$/proxy score was positive. }
\label{tbl:meansandcorr}
\end{table}

\subsubsection{Correlations Between Model and Human Scores}
\label{sec:exp2resultscorr}

We also compute the correlation between model-derived $\alpha$-scores and human-derived proxy scores, to see how model behavior aligns with human judgments; if the two do align well, we expect to see a strong, positive correlation between them.

\begin{figure}
\includegraphics[width=\linewidth]{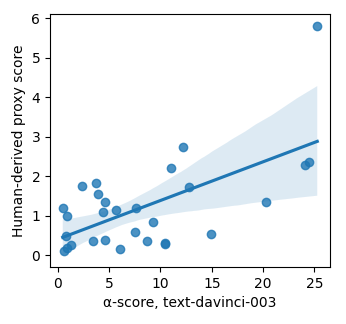}
\caption{From Experiment 2A\textemdash scatterplot of $\alpha$-scores produced by \texttt{text-davinci-003}, against human proxy scores for the same datapoints.}
\label{fig:gpt3td003corr}
\end{figure}

Table \ref{tbl:meansandcorr} shows model-wise Pearson correlation coefficients between $\alpha$-scores and human proxy scores, along with corresponding $p$-values.
Many models fail to produce correlations that are significant at $p < 0.05$; \texttt{text-davinci-003} and Llama 2 at 13B, however, both produce highly significant correlation scores.
The former also produces the highest correlation score, at around 0.62.
See Figure \ref{fig:gpt3td003corr} for a scatterplot of its $\alpha$-scores against corresponding human proxy scores; the plot lends further evidence to this correlation.

\subsubsection{Proportion of Positive $\alpha$-Scores}
Lastly, we compute the proportion of datapoints for which models produced positive $\alpha$-scores, which allows us to assess whether the models behave consistently across datapoints. Like with correlations, we observe an effect of model size: larger models perform well, with several yielding positive $\alpha$-scores for over 90\% of the data, and once again, \texttt{text-davinci-003} performs the best.

\subsection{Discussion}
\label{sec:exp2discussion}
These results suggest that a wide range of LLMs may be sensitive to the meaning ambiguity in scope ambiguous sentences.
The positive mean $\alpha$-scores provide evidence that larger or more powerful models (i.e. those besides GPT-2 small) distinguish between scope ambiguous and non-scope ambiguous sentences in a manner consistent with their meanings.
Similarly, the statistically significant correlations we see between some models' $\alpha$-scores and human proxy scores suggest that, at least for certain models, this behaviour correlates well with human judgments.
These models also produce positive $\alpha$-scores for a high proportion of the data, indicating a high level consistency in this behaviour.

Comparing chat and vanilla versions of Llama 2 also reveals an interesting pattern.
As Table \ref{tbl:meansandcorr} shows, chat versions of Llama 2 produce slightly higher mean $\alpha$-scores and correlations than their non-chat equivalents, but also lower proportions of positive $\alpha$-scores\textemdash indicating increased alignment with human judgments on several sentences, but lower overall consistency.

The broader takeaway, however, is that several LLMs appear sensitive to a meaning ambiguity that arises from the presence of different possible semantic structures, which vary vis-\`a-vis scope relations.
Consequently, although the current work does not investigate how or where models represent scope, these results suggest that LLMs capture scope-related phenomena.

\section{Re-evaluation on Expanded Datasets}
\label{sec:exp3}
The results from Experiments 1A and 2A are promising, but rely on relatively small datasets that contain 153 and 29 unique datapoints respectively\textemdash raising questions of how generalizable our results are.
As a followup, we therefore considerably expand these two datasets, and rerun the same experiments described in Sections \ref{sec:experiment1a} and \ref{sec:exp2a} on the expanded datasets.

\subsection{Dataset Expansion Process}
\paragraph{Experiment 1 Dataset Expansion}
To expand our Experiment 1A dataset, we begin by annotating it for both semantic operator and quantifier types: whether an ambiguity arose out of a negation-quantifier or quantifier-quantifier interaction, as well as whether the quantifier was an existential quantifier, universal quantifier, number or indefinite.
Combined with scope reading preference labels (see Section \ref{sec:exp1dataset}), this gave us 13 categories of scope reading and operator combinations (e.g. \texttt{negation-indefinite\_surface}, or \texttt{number-universal\_inverse}).
Following this categorization, we add manually handcrafted examples to any sparse categories, so each contains at least 10 unique datapoints.

We then use {GPT-4} to expand the dataset. 
For each category in our annotated dataset, we randomly sample 5 datapoints, and instruct GPT-4 to produce 10 novel datapoints based on them.
We repeat this process ten times, such that we have 100 datapoints generated from each of our annotated categories.
We then manually inspect the combined 1,300 generated datapoints, removing duplicates, and dropping or editing low-quality datapoints; this left us with 1,062 datapoints.
Finally, we run a crowd-sourced study via Prolific (278 participants, $\sim$5 ratings per datapoint), similar to those described in Sections \ref{sec:exp1dataset} and \ref{sec:exp1baselines}, to obtain our gold labels for preferred scope readings, and filter out any datapoints that received low inter-subject agreement.
This process eventually yields 837 unique scope-ambiguous sentences (with accompanying `options', and human preference labels); 534 receive a preferred surface scope reading, and 303 an inverse scope reading.

\paragraph{Experiment 2 Dataset Expansion}
We use a process similar to the one for Experiment 1A.

We first split our Experiment 2A dataset into categories based on whether the datapoints involve negation-quantifier, adverb-quantifier or quantifier-quantifier interactions.
We then run the same sampling, generation and manual filtering process as with the Experiment 1A dataset, giving us 126 datapoints from 300 GPT-4 generated datapoints.
Finally, we run another study via Prolific (223 participants, $\sim$8 ratings per sentence-followup pair), and use this human judgement data to further filter the dataset.
Our final dataset consists of 110 unique datapoints, where each datapoint consists of a scope-ambiguous sentence, control sentence, follow-up supporting an inverse scope reading, and follow-up supporting a surface scope reading.

\subsection{Experiment 1B}

\begin{table}
\centering
\resizebox{\columnwidth}{!}{
\begin{tabular}{@{}lcccccc@{}}
    \toprule
    \multirow{2}{*}{\textbf{Source}} & \multicolumn{2}{c}{\textbf{Accuracy}} & \multicolumn{2}{c}{\textbf{Surface Acc.}} & \multicolumn{2}{c}{\textbf{Inverse Acc.}} \\\cmidrule(lr){2-3}\cmidrule(lr){4-5}\cmidrule(lr){6-7}
    & \textit{test} & \textit{control} & \textit{test} & \textit{control} & \textit{test} & \textit{control} \\\midrule
    Llama2-7b & 0.64 & 0.64 & 0.62 & 0.63 & 0.66 & 0.65 \\
    Llama2-7b-chat & 0.57 & 0.58 & 0.58 & 0.58 & 0.56 & 0.58 \\
    Llama2-13b & 0.71 & 0.66 & 0.75 & 0.66 & 0.65 & 0.67 \\
    Llama2-13b-chat & 0.75 & 0.67 & 0.77 & 0.64 & 0.73 & 0.73 \\
    Llama2-70b & 0.89 & 0.72 & 0.91 & 0.74 & 0.84 & 0.69 \\
    Llama2-70b-chat & 0.83 & 0.65 & 0.83 & 0.64 & 0.82 & 0.67 \\
    GPT-3-davinci & 0.64 & 0.60 & 0.68 & 0.62 & 0.58 & 0.58 \\
    GPT-3.5-td002 & 0.84 & 0.70 & 0.89 & 0.69 & 0.74 & 0.72 \\
    GPT-3.5-td003 & 0.87 & 0.70 & 0.90 & 0.72 & 0.80 & 0.68 \\
    GPT-3.5-turbo & 0.79 & 0.65 & 0.86 & 0.65 & 0.68 & 0.65 \\
    \textbf{GPT-4*} & \textbf{0.96} & \textbf{0.72} & \textbf{0.97} & \textbf{0.73} & \textbf{0.93} & \textbf{0.70} \\
    \bottomrule
\end{tabular} 
}
\caption{Results from Experiment 1B: model accuracy, as well as accuracy on sentences that had a preferred surface or inverse reading. In the test setting, the ambiguous sentence is present in the prompt; in the control setting it is dropped. *Expanded dataset was also generated by \texttt{GPT-4}, albeit in a different setting and with a different system prompt.}
\label{tbl:exp3.1}
\end{table}

We re-run Experiment 1A on the expanded dataset; Table \ref{tbl:exp3.1} shows our results. 
As can be seen, the general patterns observed in Experiment 1A (see Section \ref{sec:exp1results}) continue to hold true even when the models are evaluated on a much larger dataset.
While some models either perform around chance or do not show a major accuracy drop in the control setting, models like GPT-4, \texttt{text-davinci-003} and \texttt{Llama2-70b} show both high performance (all above 85\% accuracy in the test setting, with GPT-4 achieving $\sim$96\% accuracy, albeit on data it produced in a separate context), as well as a drop-off in performance in the control setting.

\subsection{Experiment 2B}
\label{sec:exp2b}

\begin{table}
\centering
\resizebox{\columnwidth}{!}{
\begin{tabular}{@{}lccccc@{}}
    \toprule
    \multirow{2}{*}{\textbf{Source}} & \multicolumn{2}{c}{\textbf{Mean Scores}} & \multicolumn{2}{c}{\textbf{Correlations}} & \multirow{2}{*}{\textbf{\boldmath$\alpha$ > 0}} \\\cmidrule(lr){2-3}\cmidrule(lr){4-5}
    & $\alpha$/proxy score & $p$-value & R-value &  $p$-value\\\midrule
    Human & 1.34 & 5.36e-23 & 1.0 & - & 1.0 \\ 
    GPT-2-small & 1.38 & 3.78e-09 & 0.25 & 7.73e-03 & 0.80 \\
    GPT-2-med & 1.88 & 1.58e-11 & 0.29 & 1.88e-03 & 0.79 \\ 
    GPT-2-large & 1.98 & 1.38e-11 & 0.37 & 5.87e-05 & 0.76 \\ 
    GPT-2-xl & 2.87 & 7.08e-17 & 0.32 & 5.59e-04 & 0.86 \\ 
    Llama2-7b & 3.94 & 1.67e-19 & 0.37 & 7.95e-05 & 0.88 \\
    Llama2-7b-chat & 5.21 & 3.05e-20 & 0.42 & 4.38e-06 & 0.89 \\
    \textbf{Llama2-13b} & 4.31 & 5.02e-23 & 0.44 & 1.31e-06 & \textbf{0.92} \\ 
    Llama2-13b-chat & 5.12 & 4.45e-21 & 0.46 & 3.37e-07 & 0.89 \\
    Llama2-70b & 4.64 & 2.26e-22 & 0.36 & 1.32e-04 & 0.88 \\ 
    Llama2-70b-chat & 5.56 & 1.01e-20 & 0.46 & 3.65e-07 & 0.85 \\
    GPT-3-davinci & 4.16 & 2.84e-18 & 0.37 & 6.01e-05 & 0.85 \\
    GPT-3.5-td002 & 4.69 & 1.99e-20 & 0.48 & 1.51e-07 & 0.87 \\
    \textbf{GPT-3.5-td003} & \textbf{7.05} & \textbf{7.17e-22} & \textbf{0.48} & \textbf{1.09e-07} & 0.90 \\
    \bottomrule
\end{tabular}
}
\caption{Results from Experiment 2B.
\underline{Mean Scores}: Mean $\alpha$ (for models) and proxy (for humans) scores with $p$-values from paired $t$-tests ($df = 109$).
\underline{Correlations}: Pearson correlation coefficients (R-values) between each model's $\alpha$ scores and human proxy scores, with derived $p$-values ($n = 110$).
\underline{$\alpha > 0$}: Proportion of datapoints where $\alpha$/proxy score was positive.}
\label{tbl:meansandcorr2}
\end{table}

Similarly, we re-run Experiment 2A on the expanded dataset; Table \ref{tbl:meansandcorr2} shows our results.
As with Experiments 1A and 1B, the general patterns observed continue to hold for the expanded dataset.
All models still produce positive mean $\alpha$-scores.
Though not as high as on the original dataset, \texttt{text-davinci-003} once again produces the highest correlation with human data, with a R-value of roughly 0.48. 
Similarly, most models show a high level of consistency in their behaviour, producing positive $\alpha$-scores from, in the case of \texttt{text-davinci-003} and \texttt{Llama2-13b}, over 90\% of the data.\footnote{GPT-2 models also do much better on the expanding dataset. This may be because in the expansion process, we ensured that the contrast between acceptable and unacceptable sentence-followup pairs (see Figure \ref{fig:exp2dataset}) was more clear cut than in the original dataset, often coming from grammatical cues, rather than world knowledge cues. GPT-2 may recognize the former more than the latter, and thus perform better here.}
Once again, Llama 2 chat models produce higher correlations and mean scores than their vanilla counterparts, but in two out of three cases, lower proportions of positive $\alpha$-scores.

\section{Discussion}
\label{sec:discussion}
Our results, which indicate that LLMs are both proficient at choosing the scope readings preferred by most humans, and sensitive to the meaning ambiguity in scope ambiguous constructions, offer further evidence of the capacity of large language models to induce semantic structure \cite[see][]{pavlick2022}, and linguistic structure more generally \cite[see][]{linzen2021syntactic, baroni2022proper}.

On the other hand, these results contrast with closely related work by \citet{Liu2023WereAL} and \citet{stengel2023zero}, who both find that LLMs struggle to model ambiguity in zero-shot contexts.
What explains this contrast?
One possible explanation is the difference in methodologies used. 
We assess models using Q\&A- and probability-based approaches (see Sections \ref{sec:exp1method} and \ref{sec:exp2method}) that implicitly test models' access to different scope readings. 
\citet{Liu2023WereAL}, on the other hand, mostly use prompting-based approaches that elicit model responses on what an ambiguous sentence may mean or entail, and \citet{stengel2023zero} assess models in terms of their abilities to logically parse ambiguous inputs.
It is possible that LLMs \textit{implicitly} capture meaning ambiguities and human-preferred interpretations, but cannot reliably produce meta-linguistic judgments or logical translations consistent with this information.
This is would be in line with findings from \citet{hu2023prompt}, which suggest meta-linguistic prompting-based approaches may underestimate LLMs' linguistic abilities.

To test this theory, we adapt a random sample of our Experiment 2B dataset to the format \citet{Liu2023WereAL} use in their True/False evaluation of models \cite[see Section 4.2 of][]{Liu2023WereAL}.
In this format, models are prompted to answer whether it is true or false that, given one of its disambiguations, an ambiguous input \textit{may}, \textit{may not}, \textit{cannot}, or \textit{can only} mean the disambiguation.

We rerun their experiment on this subset of our data; our results, shown in Table \ref{tbl:tfeval}, are similar to the authors' findings on their own data.
Most models do poorly on this task, performing around chance (50\%), with GPT-4 and \texttt{GPT-3.5-turbo} only achieving 64\% accuracy.
As shown in Section \ref{sec:exp2b}, however, using the dataset in our experimental format yields positive results that contrast this poor performance.
This divergence highlights the importance of diverse approaches to investigating the linguistic capacities of language models; our results suggest that probability- and prompting-based methods may yield differing conclusions.

\section{Conclusion}
\label{sec:conclusion}
In this paper, we investigated how different autoregressive language models treat scope ambiguities.
In doing so, we introduced novel datasets that contain a joint total of roughly 1,000 unique and diverse scope-ambiguous sentences, annotated for human judgments\textemdash the largest of this kind.
Our results indicate that LLMs are able to exhibit behaviour in line with human preferences of interpretation\textemdash informed at least in part by background knowledge\textemdash as well as compatible with different types of semantic structures.
Finally, the contrast between our findings and those of other recent works emphasizes the need for diverse approaches in assessing the linguistic capacities of large language models. 

\begin{table}
\centering
\resizebox{\columnwidth}{!}{
\begin{tabular}{@{}lcc@{}}
    \toprule
    Model & Mean T/F Accuracy & Mean T/F Prob. Density \\
    \midrule
    Llama2-7b & 0.54 & 0.43 \\
    Llama2-7b-chat & 0.44 & 0.99 \\
    Llama2-13b & 0.51 & 0.24 \\
    Llama2-13b-chat & 0.58 & 0.99 \\
    Llama2-70b & 0.59 & 0.31 \\
    Llama2-70b-chat & 0.57 & 0.98 \\
    GPT-3.5-turbo & 0.64 & NA \\
    GPT-4 & 0.64 & NA \\
    \bottomrule
\end{tabular}
}
\caption{Results from running \citet{Liu2023WereAL}'s T/F evaluation. \underline{Mean T/F Accuracy:} Average accuracy of model's responses. \underline{Mean T/F Prob. Density:} Average probability density of the union of `True' and `False' tokens as responses given the prompt input.}
\label{tbl:tfeval}
\end{table}

\section{Limitations}
Aside from its focus only on English, one constraint of this work is that it does not assess how context affects scope reading preferences. 

\begin{exe}
\ex \label{ex:diffbgs}Ada often studies with a few of her friends.
\begin{xlist}
     \ex \label{ex:diffbgssurface}Context: Ada finds it hard to study alone, so she generally invites others for joint study sessions.
    \ex \label{ex:diffbgsinv}Context: Ada, Rohan and Jo are good friends in the same program, and prepare for exams together.
\end{xlist}
\end{exe}

(\ref{ex:diffbgs}) is ambiguous between a surface scope reading ((\ref{ex:diffbgs}) refers to no friends in particular) and an inverse scope reading ((\ref{ex:diffbgs}) refers to some specific friends). 
Different background contexts can prompt different readings: (\ref{ex:diffbgssurface}) prompts the surface scope reading, while (\ref{ex:diffbgsinv}) prompts the inverse scope reading.
Our work does not address such effects.

At a higher level, while this work shows how LLMs treat scope-ambiguous inputs, it also does not reveal how or where models represent scope.
Parallel work on model interpretability \cite[such as causal mediation analysis, e.g.,][]{vig2020investigating, finlayson2021causal, pmlr-v162-geiger22a} could provide exciting insights to this question.

\section*{Ethics Statement}

There are no obvious risks or harms associated with this experimental study.
Experiments that involved human participants were approved by our university's ethics board.
Human participants in our studies were recruited online via Prolific, and were paid on average a minimum of US\$15.00 per hour. 
Wherever anonymous participant responses were
to be made publicly accessible, participants were informed of how their responses would be used for the purposes of the study, and their rights with regard to their submitted data.

\section*{Acknowledgements}
This work was partly funded by a Doctoral Training Award from the \textit{Fonds de Recherche du Qu\'{e}bec\textemdash Soci\'{e}t\'{e} et Culture}.
We also thank Scott AnderBois, Justyna Grudzi\'{n}ska and the Law School Admission Council for access to the materials used in Experiment 1A, as well as Kristina Toutanova and the anonymous reviewers for their valuable feedback on prior versions of this work.

\bibliography{main}

\begin{thebibliography}{48}
\expandafter\ifx\csname natexlab\endcsname\relax\def\natexlab#1{#1}\fi

\bibitem[{AlKhamissi et~al.(2022)AlKhamissi, Li, Celikyilmaz, Diab, and Ghazvininejad}]{alkhamissi2022review}
Badr AlKhamissi, Millicent Li, Asli Celikyilmaz, Mona Diab, and Marjan Ghazvininejad. 2022.
\newblock \href {https://arxiv.org/abs/2204.06031} {A review on language models as knowledge bases}.
\newblock \emph{arXiv preprint arXiv:2204.06031}.

\bibitem[{AnderBois et~al.(2012)AnderBois, Brasoveanu, and Henderson}]{anderbois2012pragmatics}
Scott AnderBois, Adrian Brasoveanu, and Robert Henderson. 2012.
\newblock The pragmatics of quantifier scope: A corpus study.
\newblock In \emph{Proceedings of Sinn und Bedeutung}, volume~16, pages 15--28.

\bibitem[{Anderson(2004)}]{anderson2004structure}
Catherine Anderson. 2004.
\newblock \emph{The Structure and Real-time Comprehension of Quantifier Scope Ambiguity}.
\newblock Northwestern University.

\bibitem[{Andrew and MacCartney(2004)}]{andrew2004statistical}
Galen Andrew and Bill MacCartney. 2004.
\newblock Statistical resolution of scope ambiguity in natural language.
\newblock \emph{Unpublished manuscript}.

\bibitem[{Baroni(2022)}]{baroni2022proper}
Marco Baroni. 2022.
\newblock On the proper role of linguistically-oriented deep net analysis in linguistic theorizing.
\newblock \emph{Algebraic structures in natural language}, pages 1--16.

\bibitem[{Belinkov and Glass(2019)}]{belinkov2019analysis}
Yonatan Belinkov and James Glass. 2019.
\newblock Analysis methods in neural language processing: A survey.
\newblock \emph{Transactions of the Association for Computational Linguistics}, 7:49--72.

\bibitem[{Brown et~al.(2020)Brown, Mann, Ryder, Subbiah, Kaplan, Dhariwal, Neelakantan, Shyam, Sastry, Askell et~al.}]{brown2020language}
Tom Brown, Benjamin Mann, Nick Ryder, Melanie Subbiah, Jared~D Kaplan, Prafulla Dhariwal, Arvind Neelakantan, Pranav Shyam, Girish Sastry, Amanda Askell, et~al. 2020.
\newblock Language models are few-shot learners.
\newblock \emph{Advances in Neural Information Processing Systems}, 33:1877--1901.

\bibitem[{Cui et~al.(2022)Cui, Hershcovich, and S{\o}gaard}]{cui2022generalized}
Ruixiang Cui, Daniel Hershcovich, and Anders S{\o}gaard. 2022.
\newblock \href {https://doi.org/10.18653/v1/2022.naacl-main.359} {Generalized quantifiers as a source of error in multilingual {NLU} benchmarks}.
\newblock In \emph{Proceedings of the 2022 Conference of the North American Chapter of the Association for Computational Linguistics: Human Language Technologies}, pages 4875--4893, Seattle, United States. Association for Computational Linguistics.

\bibitem[{Dettmers et~al.(2022)Dettmers, Lewis, Belkada, and Zettlemoyer}]{dettmers2022llm}
Tim Dettmers, Mike Lewis, Younes Belkada, and Luke Zettlemoyer. 2022.
\newblock \href {https://proceedings.neurips.cc/paper_files/paper/2022/file/c3ba4962c05c49636d4c6206a97e9c8a-Paper-Conference.pdf} {{GPT3.int8()}: 8-bit matrix multiplication for transformers at scale}.
\newblock In \emph{Advances in Neural Information Processing Systems}, volume~35, pages 30318--30332. Curran Associates, Inc.

\bibitem[{Ettinger(2020)}]{ettinger2020bert}
Allyson Ettinger. 2020.
\newblock What {BERT} is not: Lessons from a new suite of psycholinguistic diagnostics for language models.
\newblock \emph{Transactions of the Association for Computational Linguistics}, 8:34--48.

\bibitem[{Ettinger et~al.(2018)Ettinger, Elgohary, Phillips, and Resnik}]{ettinger2018assessing}
Allyson Ettinger, Ahmed Elgohary, Colin Phillips, and Philip Resnik. 2018.
\newblock \href {https://aclanthology.org/C18-1152} {Assessing composition in sentence vector representations}.
\newblock In \emph{Proceedings of the 27th International Conference on Computational Linguistics}, pages 1790--1801, Santa Fe, New Mexico, USA. Association for Computational Linguistics.

\bibitem[{Finlayson et~al.(2021)Finlayson, Mueller, Gehrmann, Shieber, Linzen, and Belinkov}]{finlayson2021causal}
Matthew Finlayson, Aaron Mueller, Sebastian Gehrmann, Stuart Shieber, Tal Linzen, and Yonatan Belinkov. 2021.
\newblock \href {https://doi.org/10.18653/v1/2021.acl-long.144} {Causal analysis of syntactic agreement mechanisms in neural language models}.
\newblock In \emph{Proceedings of the 59th Annual Meeting of the Association for Computational Linguistics and the 11th International Joint Conference on Natural Language Processing (Volume 1: Long Papers)}, pages 1828--1843, Online. Association for Computational Linguistics.

\bibitem[{Futrell et~al.(2019)Futrell, Wilcox, Morita, Qian, Ballesteros, and Levy}]{futrell-etal-2019-neural}
Richard Futrell, Ethan Wilcox, Takashi Morita, Peng Qian, Miguel Ballesteros, and Roger Levy. 2019.
\newblock \href {https://doi.org/10.18653/v1/N19-1004} {Neural language models as psycholinguistic subjects: Representations of syntactic state}.
\newblock In \emph{Proceedings of the 2019 Conference of the North {A}merican Chapter of the Association for Computational Linguistics: Human Language Technologies, Volume 1 (Long and Short Papers)}, pages 32--42, Minneapolis, Minnesota. Association for Computational Linguistics.

\bibitem[{Geiger et~al.(2022)Geiger, Wu, Lu, Rozner, Kreiss, Icard, Goodman, and Potts}]{pmlr-v162-geiger22a}
Atticus Geiger, Zhengxuan Wu, Hanson Lu, Josh Rozner, Elisa Kreiss, Thomas Icard, Noah Goodman, and Christopher Potts. 2022.
\newblock \href {https://proceedings.mlr.press/v162/geiger22a.html} {Inducing causal structure for interpretable neural networks}.
\newblock In \emph{Proceedings of the 39th International Conference on Machine Learning}, volume 162 of \emph{Proceedings of Machine Learning Research}, pages 7324--7338. PMLR.

\bibitem[{Heinzerling and Inui(2021)}]{heinzerling2021language}
Benjamin Heinzerling and Kentaro Inui. 2021.
\newblock \href {https://doi.org/10.18653/v1/2021.eacl-main.153} {Language models as knowledge bases: On entity representations, storage capacity, and paraphrased queries}.
\newblock In \emph{Proceedings of the 16th Conference of the European Chapter of the Association for Computational Linguistics: Main Volume}, pages 1772--1791, Online. Association for Computational Linguistics.

\bibitem[{Hewitt and Manning(2019)}]{hewitt2019structural}
John Hewitt and Christopher~D Manning. 2019.
\newblock A structural probe for finding syntax in word representations.
\newblock In \emph{Proceedings of the 2019 Conference of the North American Chapter of the Association for Computational Linguistics: Human Language Technologies, Volume 1 (Long and Short Papers)}, pages 4129--4138.

\bibitem[{Higgins and Sadock(2003)}]{higgins2003machine}
Derrick Higgins and Jerrold~M Sadock. 2003.
\newblock A machine learning approach to modeling scope preferences.
\newblock \emph{Computational Linguistics}, 29(1):73--96.

\bibitem[{Hu and Levy(2023)}]{hu2023prompt}
Jennifer Hu and Roger Levy. 2023.
\newblock \href {https://arxiv.org/abs/2305.13264} {Prompt-based methods may underestimate large language models' linguistic generalizations}.
\newblock \emph{arXiv preprint arXiv:2305.13264}.

\bibitem[{Jang and Lukasiewicz(2023)}]{jang2023consistency}
Myeongjun Jang and Thomas Lukasiewicz. 2023.
\newblock \href {https://arxiv.org/abs/2303.06273} {Consistency analysis of {ChatGPT}}.
\newblock \emph{arXiv preprint arXiv:2303.06273}.

\bibitem[{Jawahar et~al.(2019)Jawahar, Sagot, and Seddah}]{jawahar2019does}
Ganesh Jawahar, Beno{\^\i}t Sagot, and Djam{\'e} Seddah. 2019.
\newblock What does {BERT} learn about the structure of language?
\newblock In \emph{ACL 2019-57th Annual Meeting of the Association for Computational Linguistics}.

\bibitem[{Kim et~al.(2019)Kim, Patel, Poliak, Xia, Wang, McCoy, Tenney, Ross, Linzen, Van~Durme, Bowman, and Pavlick}]{kim2019probing}
Najoung Kim, Roma Patel, Adam Poliak, Patrick Xia, Alex Wang, Tom McCoy, Ian Tenney, Alexis Ross, Tal Linzen, Benjamin Van~Durme, Samuel~R. Bowman, and Ellie Pavlick. 2019.
\newblock \href {https://doi.org/10.18653/v1/S19-1026} {Probing what different {NLP} tasks teach machines about function word comprehension}.
\newblock In \emph{Proceedings of the Eighth Joint Conference on Lexical and Computational Semantics (*{SEM} 2019)}, pages 235--249, Minneapolis, Minnesota. Association for Computational Linguistics.

\bibitem[{Kurtzman and MacDonald(1993)}]{kurtzman1993resolution}
Howard~S Kurtzman and Maryellen~C MacDonald. 1993.
\newblock Resolution of quantifier scope ambiguities.
\newblock \emph{Cognition}, 48(3):243--279.

\bibitem[{Leczkowski et~al.(2022)Leczkowski, Grudzi{\'n}ska, Guzm{\'a}n, Wawer, and Siemieniuk}]{leczkowski2022prepositions}
Aleksander Leczkowski, Justyna Grudzi{\'n}ska, Manuel~Vargas Guzm{\'a}n, Aleksander Wawer, and Aleksandra Siemieniuk. 2022.
\newblock Prepositions matter in quantifier scope disambiguation.
\newblock In \emph{Proceedings of the 29th International Conference on Computational Linguistics}, pages 3960--3970.

\bibitem[{Linzen and Baroni(2021)}]{linzen2021syntactic}
Tal Linzen and Marco Baroni. 2021.
\newblock Syntactic structure from deep learning.
\newblock \emph{Annual Review of Linguistics}, 7:195--212.

\bibitem[{Linzen et~al.(2016)Linzen, Dupoux, and Goldberg}]{linzen2016assessing}
Tal Linzen, Emmanuel Dupoux, and Yoav Goldberg. 2016.
\newblock Assessing the ability of {LSTM}s to learn syntax-sensitive dependencies.
\newblock \emph{Transactions of the Association for Computational Linguistics}, 4:521--535.

\bibitem[{Liu et~al.(2023)Liu, Wu, Michael, Suhr, West, Koller, Swayamdipta, Smith, and Choi}]{Liu2023WereAL}
Alisa Liu, Zhaofeng Wu, Julian Michael, Alane Suhr, Peter West, Alexander Koller, Swabha Swayamdipta, Noah~A. Smith, and Yejin Choi. 2023.
\newblock \href {https://arxiv.org/abs/2304.14399} {We're afraid language models aren't modeling ambiguity}.
\newblock \emph{arXiv preprint arXiv:2304.14399}.

\bibitem[{Liu et~al.(2019)Liu, Ott, Goyal, Du, Joshi, Chen, Levy, Lewis, Zettlemoyer, and Stoyanov}]{liu2019roberta}
Yinhan Liu, Myle Ott, Naman Goyal, Jingfei Du, Mandar Joshi, Danqi Chen, Omer Levy, Mike Lewis, Luke Zettlemoyer, and Veselin Stoyanov. 2019.
\newblock \href {https://arxiv.org/abs/1907.11692} {{RoBERTa}: A robustly optimized {BERT} pretraining approach}.
\newblock \emph{arXiv preprint arXiv:1907.11692}.

\bibitem[{Manshadi and Allen(2011)}]{manshadi2011unrestricted}
Mehdi Manshadi and James Allen. 2011.
\newblock Unrestricted quantifier scope disambiguation.
\newblock In \emph{Proceedings of TextGraphs-6: Graph-based Methods for Natural Language Processing}, pages 51--59.

\bibitem[{Misra(2022)}]{misra2022minicons}
Kanishka Misra. 2022.
\newblock \href {https://arxiv.org/abs/2203.13112} {minicons: Enabling flexible behavioral and representational analyses of transformer language models}.
\newblock \emph{arXiv preprint arXiv:2203.13112}.

\bibitem[{OpenAI(2023)}]{openai2023gpt4}
OpenAI. 2023.
\newblock \href {http://arxiv.org/abs/2303.08774} {{GPT-4} technical report}.

\bibitem[{Ouyang et~al.(2022)Ouyang, Wu, Jiang, Almeida, Wainwright, Mishkin, Zhang, Agarwal, Slama, Ray et~al.}]{ouyang2022training}
Long Ouyang, Jeffrey Wu, Xu~Jiang, Diogo Almeida, Carroll Wainwright, Pamela Mishkin, Chong Zhang, Sandhini Agarwal, Katarina Slama, Alex Ray, et~al. 2022.
\newblock Training language models to follow instructions with human feedback.
\newblock \emph{Advances in Neural Information Processing Systems}, 35:27730--27744.

\bibitem[{Pavlick(2022)}]{pavlick2022}
Ellie Pavlick. 2022.
\newblock \href {https://doi.org/10.1146/annurev-linguistics-031120-122924} {Semantic structure in deep learning}.
\newblock \emph{Annual Review of Linguistics}, 8(1):447--471.

\bibitem[{Radford et~al.(2019)Radford, Wu, Child, Luan, Amodei, Sutskever et~al.}]{radford2019language}
Alec Radford, Jeffrey Wu, Rewon Child, David Luan, Dario Amodei, Ilya Sutskever, et~al. 2019.
\newblock Language models are unsupervised multitask learners.
\newblock \emph{OpenAI blog}, 1(8):9.

\bibitem[{Rasmussen(2022)}]{rasmussen2022broad}
Nathan~Ellis Rasmussen. 2022.
\newblock \emph{Broad-domain Quantifier Scoping with RoBERTa}.
\newblock Ph.D. thesis, The Ohio State University.

\bibitem[{Richardson et~al.(2020)Richardson, Hu, Moss, and Sabharwal}]{richardson2020probing}
Kyle Richardson, Hai Hu, Lawrence Moss, and Ashish Sabharwal. 2020.
\newblock Probing natural language inference models through semantic fragments.
\newblock In \emph{Proceedings of the AAAI Conference on Artificial Intelligence}, volume~34, pages 8713--8721.

\bibitem[{Roberts et~al.(2020)Roberts, Raffel, and Shazeer}]{roberts2020much}
Adam Roberts, Colin Raffel, and Noam Shazeer. 2020.
\newblock \href {https://doi.org/10.18653/v1/2020.emnlp-main.437} {How much knowledge can you pack into the parameters of a language model?}
\newblock In \emph{Proceedings of the 2020 Conference on Empirical Methods in Natural Language Processing (EMNLP)}, pages 5418--5426, Online. Association for Computational Linguistics.

\bibitem[{Saba and Corriveau(2001)}]{saba2001plausible}
Walid~S Saba and Jean-Pierre Corriveau. 2001.
\newblock Plausible reasoning and the resolution of quantifier scope ambiguities.
\newblock \emph{Studia Logica}, 67:271--289.

\bibitem[{Schuster and Linzen(2022)}]{schuster-linzen-2022-sentence}
Sebastian Schuster and Tal Linzen. 2022.
\newblock \href {https://doi.org/10.18653/v1/2022.naacl-main.71} {When a sentence does not introduce a discourse entity, transformer-based models still sometimes refer to it}.
\newblock In \emph{Proceedings of the 2022 Conference of the North American Chapter of the Association for Computational Linguistics: Human Language Technologies}, pages 969--982, Seattle, United States. Association for Computational Linguistics.

\bibitem[{Shwartz and Dagan(2019)}]{shwartz2019}
Vered Shwartz and Ido Dagan. 2019.
\newblock Still a pain in the neck: Evaluating text representations on lexical composition.
\newblock \emph{Transactions of the Association for Computational Linguistics}, 7:403--419.

\bibitem[{Stengel-Eskin et~al.(2023)Stengel-Eskin, Rawlins, and Van~Durme}]{stengel2023zero}
Elias Stengel-Eskin, Kyle Rawlins, and Benjamin Van~Durme. 2023.
\newblock \href {https://arxiv.org/abs/2306.00824} {Zero and few-shot semantic parsing with ambiguous inputs}.
\newblock \emph{arXiv preprint arXiv:2306.00824}.

\bibitem[{Touvron et~al.(2023)Touvron, Martin, Stone, Albert, Almahairi, Babaei, Bashlykov, Batra, Bhargava, Bhosale et~al.}]{touvron2023llama}
Hugo Touvron, Louis Martin, Kevin Stone, Peter Albert, Amjad Almahairi, Yasmine Babaei, Nikolay Bashlykov, Soumya Batra, Prajjwal Bhargava, Shruti Bhosale, et~al. 2023.
\newblock \href {https://arxiv.org/abs/2307.09288} {Llama 2: Open foundation and fine-tuned chat models}.
\newblock \emph{arXiv preprint arXiv:2307.09288}.

\bibitem[{Tsiolis(2020)}]{tsiolis2020quantifier}
KC~Tsiolis. 2020.
\newblock Quantifier scope disambiguation.
\newblock \emph{Unpublished manuscript}.

\bibitem[{Vig et~al.(2020)Vig, Gehrmann, Belinkov, Qian, Nevo, Singer, and Shieber}]{vig2020investigating}
Jesse Vig, Sebastian Gehrmann, Yonatan Belinkov, Sharon Qian, Daniel Nevo, Yaron Singer, and Stuart Shieber. 2020.
\newblock Investigating gender bias in language models using causal mediation analysis.
\newblock \emph{Advances in Neural Information Processing Systems}, 33:12388--12401.

\bibitem[{Wijnholds(2023)}]{wijnholds2023assessing}
Gijs Wijnholds. 2023.
\newblock Assessing monotonicity reasoning in {Dutch} through natural language inference.
\newblock In \emph{Findings of the Association for Computational Linguistics: EACL 2023}, pages 1464--1470.

\bibitem[{Yanaka et~al.(2020)Yanaka, Mineshima, Bekki, and Inui}]{yanaka2020neural}
Hitomi Yanaka, Koji Mineshima, Daisuke Bekki, and Kentaro Inui. 2020.
\newblock \href {https://doi.org/10.18653/v1/2020.acl-main.543} {Do neural models learn systematicity of monotonicity inference in natural language?}
\newblock In \emph{Proceedings of the 58th Annual Meeting of the Association for Computational Linguistics}, pages 6105--6117, Online. Association for Computational Linguistics.

\bibitem[{Yanaka et~al.(2019)Yanaka, Mineshima, Bekki, Inui, Sekine, Abzianidze, and Bos}]{yanaka2019can}
Hitomi Yanaka, Koji Mineshima, Daisuke Bekki, Kentaro Inui, Satoshi Sekine, Lasha Abzianidze, and Johan Bos. 2019.
\newblock \href {https://doi.org/10.18653/v1/W19-4804} {Can neural networks understand monotonicity reasoning?}
\newblock In \emph{Proceedings of the 2019 ACL Workshop BlackboxNLP: Analyzing and Interpreting Neural Networks for NLP}, pages 31--40, Florence, Italy. Association for Computational Linguistics.

\bibitem[{Yu and Ettinger(2020)}]{yu2020assessing}
Lang Yu and Allyson Ettinger. 2020.
\newblock \href {https://doi.org/10.18653/v1/2020.emnlp-main.397} {Assessing phrasal representation and composition in transformers}.
\newblock In \emph{Proceedings of the 2020 Conference on Empirical Methods in Natural Language Processing (EMNLP)}, pages 4896--4907, Online. Association for Computational Linguistics.

\bibitem[{Yu and Ettinger(2021)}]{yu2021interplay}
Lang Yu and Allyson Ettinger. 2021.
\newblock \href {https://doi.org/10.18653/v1/2021.findings-acl.201} {On the interplay between fine-tuning and composition in transformers}.
\newblock In \emph{Findings of the Association for Computational Linguistics: ACL-IJCNLP 2021}, pages 2279--2293, Online. Association for Computational Linguistics.

\end{thebibliography}
\bibliographystyle{acl_natbib}

\end{document}